*Article*

# On the Role of Field of View for Occlusion Removal with Airborne Optical Sectioning

**Francis Seits, Indrajit Kurmi, Rakesh John Amala Arokia Nathan, Rudolf Ortner, and Oliver Bimber\***

Johannes Kepler University Linz
**\*Correspondence:** oliver.bimber@jku.at; Tel.: +43-732-2468-6631

**Abstract:** Occlusion caused by vegetation is an essential problem for remote sensing applications in areas, such as search and rescue, wildfire detection, wildlife observation, surveillance, border control, and others. Airborne Optical Sectioning (AOS) is an optical, wavelength-independent synthetic aperture imaging technique that supports computational occlusion removal in real-time. It can be applied with manned or unmanned aircrafts, such as drones. In this article, we demonstrate a relationship between forest density and field of view (FOV) of applied imaging systems. This finding was made with the help of a simulated procedural forest model which offers the consideration of more realistic occlusion properties than our previous statistical model. While AOS has been explored with automatic and autonomous research prototypes in the past, we present a free AOS integration for DJI systems. It enables bluelight organizations and others to use and explore AOS with compatible, manually operated, off-the-shelf drones. The (digitally cropped) default FOV for this implementation was chosen based on our new finding.

## 1. Introduction

Synthetic apertures approximate the signal of a single hypothetical wide aperture sensor with either an array of static small aperture sensors or a single moving small aperture sensor whose individual signals are computationally combined to increase resolution, depth-of-field, frame rate, contrast, and signal-to-noise ratio. This principle has been used in many fields, such as for synthetic aperture radar [1-3], synthetic aperture radio telescopes [4-5], interferometric synthetic aperture microscopy [6], synthetic aperture sonar [7-8], synthetic aperture ultrasound [9-10], and synthetic aperture LiDAR / synthetic aperture imaging laser [11-12]. In the visible range, synthetic aperture imaging [13-20] has been used together with large camera arrays that capture structured light fields (regularly sampled multiscopic scene representations) and enable the computation of virtual views with maximal synthetic apertures that correspond to the physical size of the applied camera array.

With Airborne Optical Sectioning (AOS) [21-31] we have introduced an optical synthetic aperture imaging technique that computationally removes occlusion caused by dense vegetation, such as forest, in real-time. It utilizes manually, automatically [21-28,30], or fully autonomously [29] operated camera drones that sample multispectral (RGB and thermal) images within a certain (synthetic aperture) area above forest (cf. Fig. 1a,b). These images are geo-registered to a digital elevation model of the ground surface using the drone's sensor data, and are then averaged to a single integral image (cf. Figs. 1d,f). These integral images reveal focused details of registered targets on the ground, while unregistered occluders above the ground, such as trunks, branches or leaves disappear in strong defocus. In principle, these integral images mimic the optical signal that an ultra-large lens would capture which spans the entire synthetic aperture area (cf. Fig. 1b). Such a lens would cause an extremely shallow depth of field – emphasizing the signal of focused targets while suppressing the signal of defocused occluders. This principle has been explored in fields where occlusion is critical, such as search and rescue [28-29] and wildlife observation [25], and has potential in many other areas, such as wildfire detection, surveillance, and border control.

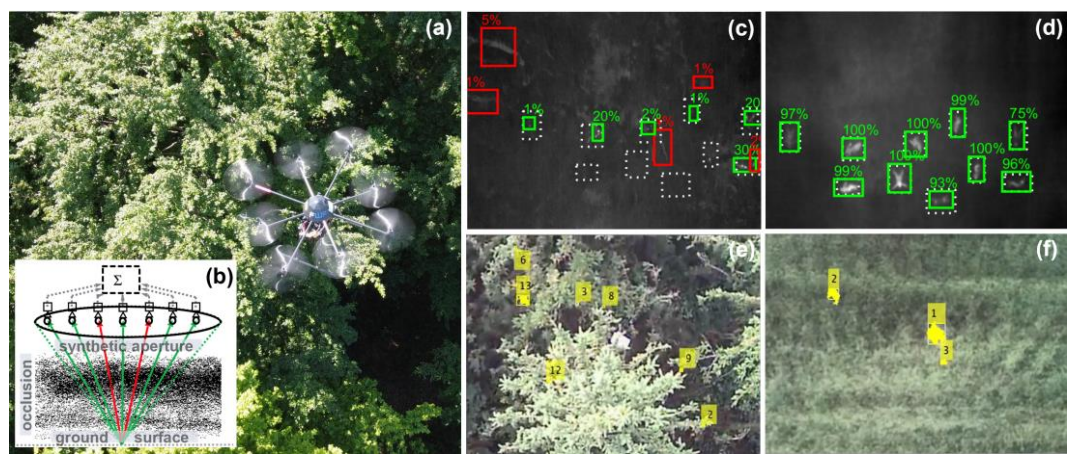

**Figure 1.** A fully autonomous drone [29] (a) during a search and rescue mission above densely occluded forest applying the AOS principle (b) for occlusion removal. Regular thermal (c) and RGB (e) aerial images suffer from occlusion, while corresponding AOS integral images (d,f) benefit from computational occlusion removal. Examples (c,d) show classifications of 10 static persons in thermal images using deep neural networks [28] (numbers are confidence scores), and examples (e,f) illustrate tracking of 3 moving persons with color anomaly detection in RGB images [31] (numbers are target IDs). While many false positives are detected in conventional aerial images (c,e), strong true positives are found in integral images (d,f).

The main advantages of AOS, when compared to alternatives, such as LiDAR or synthetic aperture radar, is its wavelength independence (same technique for near/far infrared and visible spectrum – addressing a wide range of applications) and its real-time capability (which is essential for many areas, such search and rescue). We have proven that common image processing tasks, like as person classification with deep neural networks [28-29] (cf. Figs. 1c,d) or color anomaly detection [31] (cf. Figs. 1e,f) perform significantly better when applied to AOS integral images compared to conventional aerial images.

Based on a statistical model that assumes uniformly sized and distributed occluders of known uniform densities, we have explained mathematically the limits of AOS and it's optimal sampling parameters [23,30]. The key findings of this statistical approach can be summarized as follows:

1. Visibility improves with the number of integrated image samples. However, there exists a limit that depends on the density of the occluder volume. An occlusion reduction beyond this limit cannot be achieved – even not with larger number of integrated image samples.
2. For a given flying speed, higher flight altitudes have no effect on scanning time but lead to smaller disparities of projected occluders (occlusion removal is inefficient and even impossible for too small disparities). Flight altitudes should therefore be chosen to be as low as possible (just above the tree crowns + safety margin).

3. A larger field of view (FOV) of the image optics reduces the efficiency of occlusion removal since more oblique viewing angles into the occlusion volume will cause more projected occlusion over longer viewing distances. Consequently, the FOV should ideally be as narrow as possible, while the loss of visibility gain caused by larger field of views might be tolerable up to a certain degree.

Especially findings 2 and 3 (i.e., flying low while imaging with small FOV optics) imply a limited coverage on the surface ground per single recording. This reduces also the overlap of multiple integrated images, and with this, the number of measurements of same surface points in different directions. A sufficiently high number of directional measurements per surface point, which makes occlusion removal efficient, must then be achieved through a larger number of sampling steps at smaller distances.

Scanning large areas with aerial imaging often employs high-flying aircrafts with wide FOV optics (e.g., wide-FOV Nadir cameras [32-35]). This maximizes ground coverage per single recording, but occlusion cannot be handled. We have already explained (finding 2) that higher flight altitudes is not of advantage for AOS scanning time, but low altitudes are beneficial for occlusion removal due to larger disparities.

In this article we want to reconsider the role the FOV for occlusion removal caused by larger overlaps of integrated images. Our statistical model [23,30] is clearly limited in the assumption of uniform occlude sizes, distribution, and density. However, this does not apply in practice for real forests, where we observe clear changes from crown tops to the surface. While the tree crowns consist of many small and dense occluders, the tree trunks near the surface are large, sparse, and few.

With the help of procedural forest simulation, we make the following contributions: We confirm findings 1 and 2 of our statistical model, but disprove finding 3. The optimal FOV is not as narrow as possible, but depends on tree and forest parameters. This finding could not be made with our simplified statistical model that assumes uniformity.

## 2. Materials and Methods

For revisiting findings 1-3, a simulated forest environment was used (cf. Fig. 2). It provides computed areal image data similar to drone footage obtained from flights over real forest. The main advantage of this approach is that several sampling parameters can be tested in a considerably short period of time and can be compared with ground-truth information. In contrast to our statistical model, this simulated model provides a more realistic basis for sampling parameter analysis.

The forest simulation[1] was realized with a procedural tree algorithm called ProcTree[2] and was implemented with WebGL. It computes 512x512px aerial images for drone flights over a predefined area and for defined sampling parameters (e.g., drone position, altitude, sampling distance, and camera FOV). The virtual rendering camera applies perspective projection and is aligned with its look-at vector parallel to the ground surface normal (i.e., pointing downwards). The simulated images were binarized to separate between unoccluded points on the ground (1, white) and occluders above the ground (0, black).

Procedural tree parameters, such as tree height (20m-25m), trunk length (4m-8m), trunk radius (20cm-50cm), and leaf size (5cm-20cm) were used to generate tree species as a mixture of conifer and broadleaf trees (cf. Fig. 2a). A seeded random generator was finally applied to generate varying trees at defined densities and degrees of similarity in repeated experiments.

---

[1] Online Executable: https://aos.tensorware.app/ , Source Code: https://github.com/tensorware/aos-simulation

[2] Source Code: https://github.com/supereggbert/proctree.js

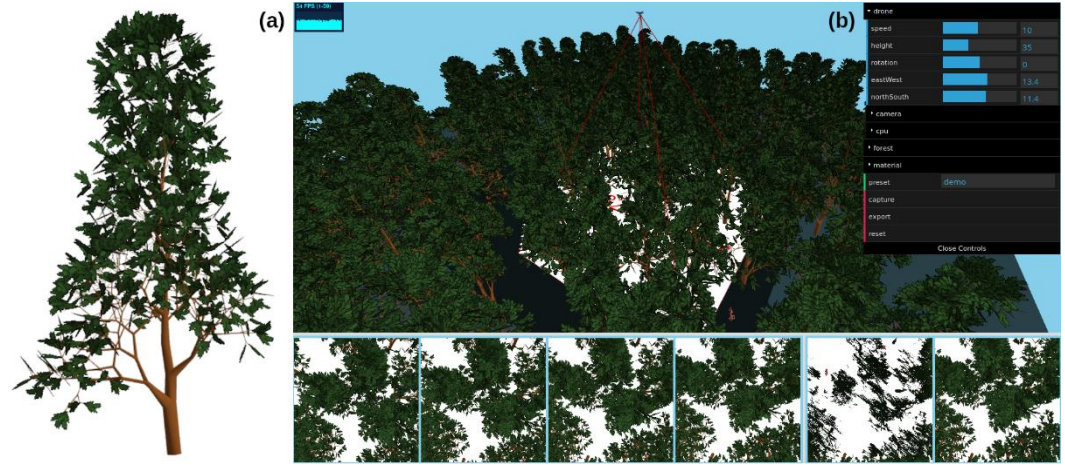

**Figure 2.** A procedural algorithm computes trees (a) for a simulated forest environment (b). Areal image data (b, bottom) is computationally generated by simulating drone flights at defined sampling parameters. Procedural trees and forest densities differ due randomness involved, but have basic structural features in common.

Environment properties such as tree species, foliage and time of year are assumed to be constant, as we are mainly interested in effects caused by changing sampling parameters. Besides sampling and procedural trees parameters, forest density is considered in three categories: sparse (133 trees/ha, cf. Fig. 3a), medium (266 trees/ha, cf. Fig. 3b), and dense (400 trees/ha, cf. Fig. 3c).

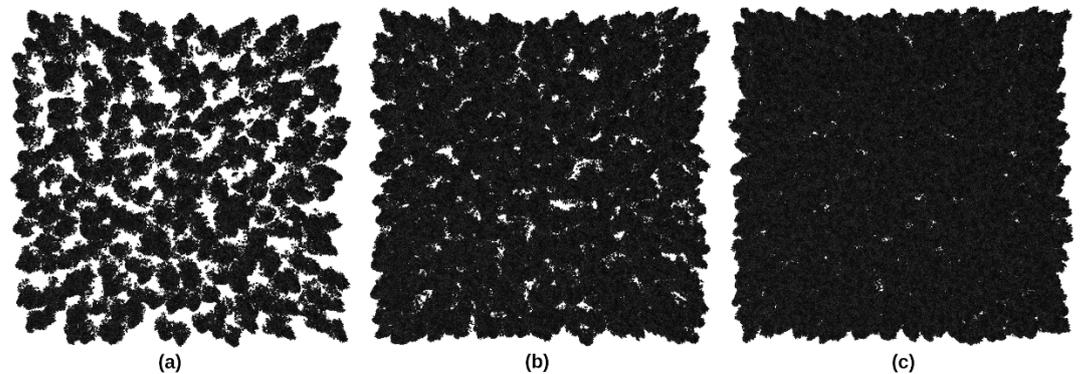

**Figure 3.** Binarized aerial views of a simulated 150x150m forest patch with different densities: 133 trees/ha (a), 266 trees/ha (b), and 400 trees/ha (c). Note, that while white pixels represent the unoccluded forest ground, black pixels are occluders. Consequently, the ratio of white to black pixels corresponds to density.

To simulate AOS, the binarized aerial images are integrated as explained in Section 1. The average grayscale values of the resulting integrals correspond to remaining occlusion (a high value represents high visibility / low occlusion, a low value represents low visibility / high occlusion).

## 3. Results

The camera's field ($FOV$ [°]) and the drone's altitude ($h$ [m]) have a direct influence on occlusion removal, as we will demonstrate below. In the further course of this article, we will use the following equations [30], which are mainly defined for 1D, but retain their meaning in 2D and 3D environments.

We define the camera's maximal possible viewing angle as

$$\alpha_{max} = \frac{FOV}{2} < 90°. \quad \text{(Eqn. 1)}$$

The camera's coverage on the ground depends on the flying altitude and the maximal possible viewing angle:

$$c = 2 \cdot h \cdot tan(\alpha_{max}). \quad \text{(Eqn. 2)}$$

How often the same point on the ground is captured from multiple drone poses is given by:

$$n = \frac{c}{d}, \quad \text{(Eqn. 3)}$$

where $d$ [m] is the sampling distance between consecutive drone positions.

Without considering occlusion, the coverage ($c$) and sampling distance ($d$) already have a direct influence on the number of times ($n$) a point on the ground could theoretically be detected.

Figure 4 illustrates simulation results for the visibility achieved with AOS at different fields of view ($FOV$=20-90°), flying altitudes ($h$=30, 40, and 50m AGL), forest densities (sparse, medium, dense), and sampling distances ($d$=0.5, 1, 1.5 and 2m). The markers indicate maximal visibility for all cases. It can clearly be seen, that lower flight altitude and smaller sampling distances (and with this, larger numbers of integrated samples for a given field of view) is always best. This confirms finding 2 and –partially– finding 1 (increasing visibility with increasing samples integrated). However, the optimal field of view always depends on forest density (which disproves finding 3): Higher densities require narrower fields of view, while wide fields of view are better for sparser forests. The reason for this finding will be discussed in Section 4.

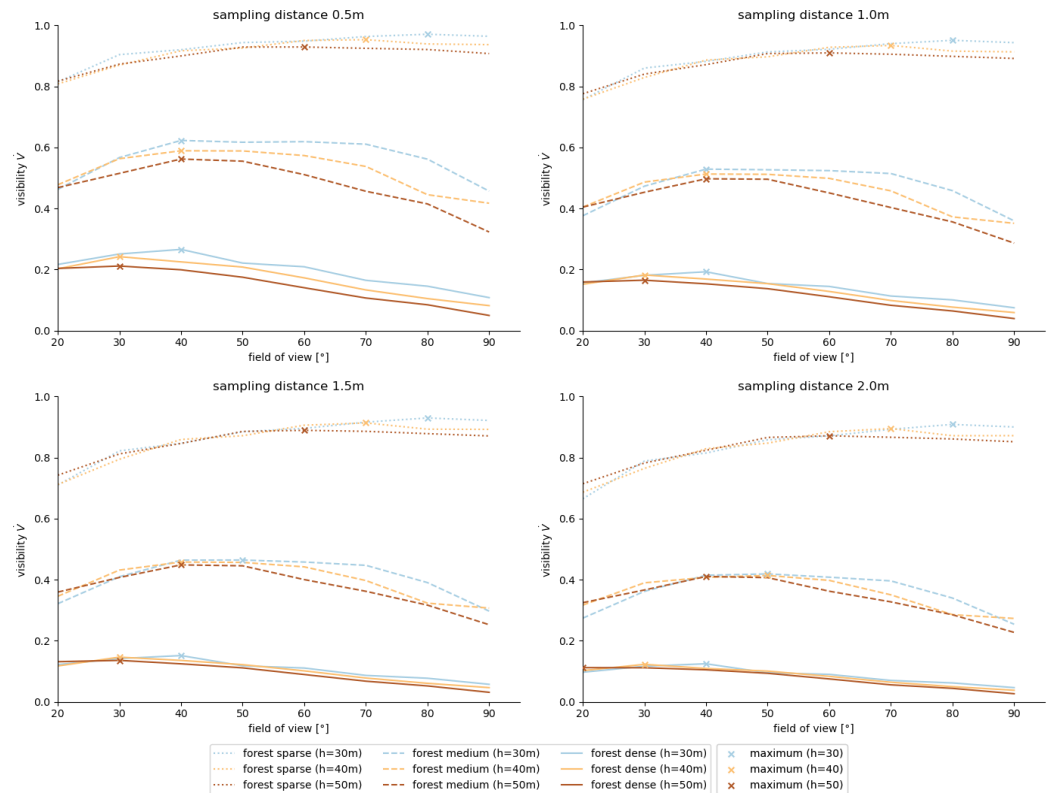

**Figure 4.** Simulated visibility achieved with AOS for varying fields of view (*FOV*), flying altitudes (*h*), sampling distances (*d*), and forest densities. While low flying altitudes and small sampling distances (i.e., large number of integrated samples) are always beneficial, the optimal field of view depends on forest density. While wide fields of view are better for sparse forests, narrow ones are better for dense forests. The markers indicate the maximum visibilities.

Figure 5 presents the same results of Fig. 4 from a different perspective. It plots (for a constant flying altitude, *h*=30m) visibility per number (*n*) of integrated samples of the same surface point. Note, that *n*, *FOV*, *d*, and ground coverage (*c*) directly correlate even without occlusion, as explained with Eqns. 1-3. Clearly, the (density-depended) upper limit of visibility gain over an increasing number of integrated samples can be seen. Higher visibility cannot be achieved as the curves flatten – even if more samples are integrated. This confirms finding 1 (limited visibility gain that depends on forest density). Visibility values as shown in Fig. 4 are highlighted in Fig. 5 with markers for each field of view, sampling distance and forest density. Figure 5 also illustrate that smaller sampling distances (and consequently larger numbers of samples *n* for a given *FOV*) is always beneficial.

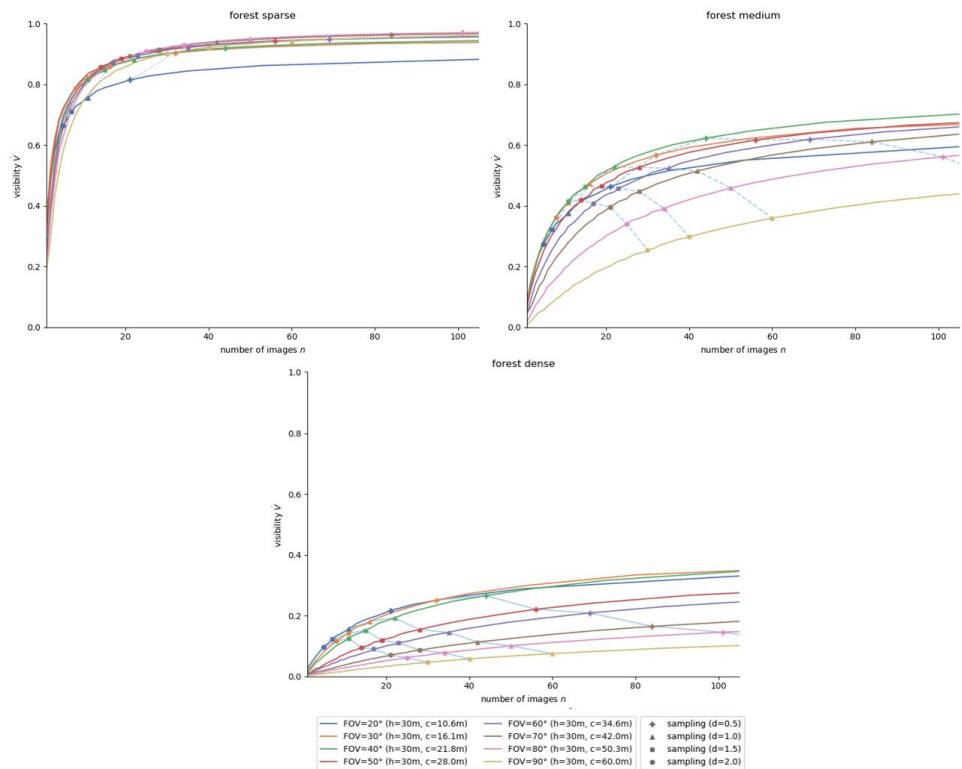

**Figure 5.** Simulated visibility achieved with AOS for varying numbers of integrated samples (*n*) and fields of view (*FOV*) at constant flying altitude (*h*=30m). The markers indicate the visibilities for different sampling distances (*d*), as shown in Fig. 4. Corresponding ground coverage (*c*) is also presented. Smaller sampling distances (and consequently larger numbers of samples *n* for a given *FOV*) is always beneficial. Flattening of curves, however, indicate the upper limit of visibility gain over an increasing number of integrated samples.

## 4. Discussion

The results presented in Section 3 confirm, for simulated procedural forests, our previous findings made with a simplified statistical model: Visibility improves with the number of integrated image samples (i.e., smaller sampling distances) and lower flying altitudes, but there exists an upper limit for occlusion removal that depends on forest density.

An essential difference between the new procedural approach and the old statistical one is the uniformity assumption of the occlusion volume (occluder size and distribution), which was considered to be constant in the latter case. For an occlusion volume of uniformly distributed and sized occluders, more oblique viewing angles will cause more projected occlusion over longer viewing distances. Consequently, such a model suggests that the *FOV* should ideally be as narrow as possible. This, however, does not hold true for conventional forests where both, occluder size and distribution, varies strongly with height.

In a simplified view, we can separate forest into two main parts (cf. Fig. 6): A trunk part that mainly consists of a relatively sparse volume of larger occluders (trunks), and a crown part that is a mainly dense volume of smaller occluders (branches and foliage).

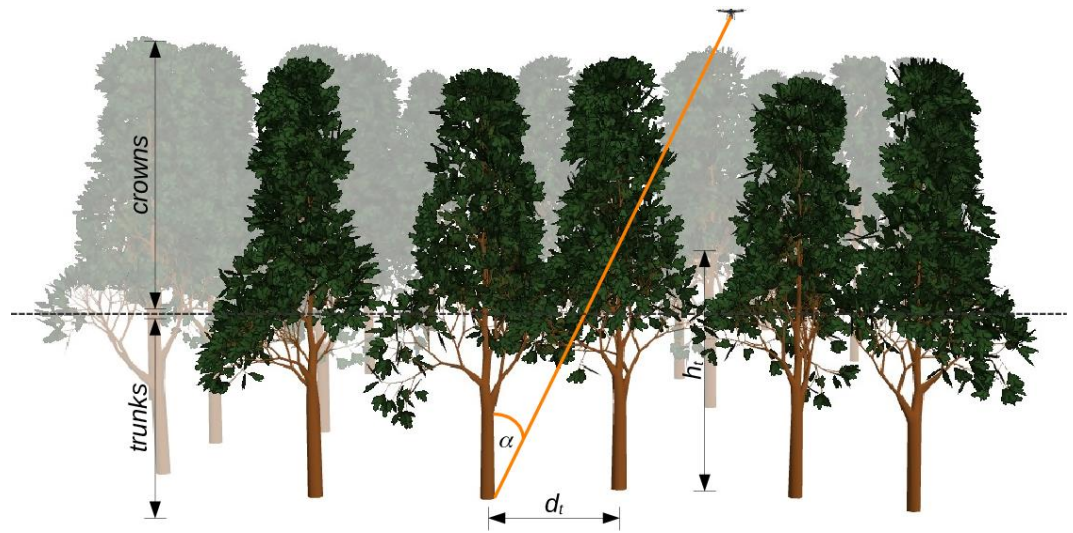

**Figure 6.** Separating forest into trunk and crown parts splits the occlusion volume into denser regions of smaller occluders (crowns), and sparser regions of larger occluders (trunks). The optimal viewing angle $\alpha$ is defined by the height of the trunk part ($h_t$) and the distance between the trees ($d_t$).

If we assume little occlusion, a large *FOV* is beneficial because (regarding to Eqns. 1-3) it maximizes ground coverage (*c*) and sampling (*n*). For dense occlusion, a small *FOV* is of advantage because it reduces projected occlusion over shorter lines of sight. In case of our simplified two-parted forest model, the optimal viewing angle $\alpha$ is defined by the height of the trunk part and the distance between the trees (Fig. 6). A smaller $\alpha$ would reduce coverage (*c*) and sampling (*n*), while a larger $\alpha$ would cause more projected occlusion (mainly of large trunks) while not increasing ground coverage any further (due to occlusion with trunks). This suggests that the optimal *FOV* for occlusion removal is $2\alpha$, or

$$FOV = 2 \cdot arctan\left(\frac{d_t}{h_t}\right), \quad \text{(Eqn. 4)}$$

where $h_t$ is the average height of the trunk part and $d_t$ the average tree distance.

For example, an average $h_t$=7m and $d_t$=3, 3.5, 4.9m (as chosen for our sparse, medium, and dense forest simulations) lead to an optimal *FOV* of around 46, 53, 70° (this tendency is, in principle, also suggested by the simulation results in Fig. 4, with a +/- 10% tolerance that can be contributed to averaging).

## 5. Conclusions

Occlusion caused by vegetation is an essential problem for remote sensing applications in areas, such as search and rescue, wildfire detection, wildlife observation, surveillance, border control, and others. Airborne Optical Sectioning (AOS) is an optical, wavelength-independent synthetic aperture imaging technique that supports computational occlusion removal in real-time. In this article, we demonstrate the relationship between forest density and optimal FOV of applied camera systems. This finding was made with the help of a simulated procedural forest model which offers the consideration of more realistic occlusion properties than our previous statistical model. If average forest properties are known, the optimal FOV can be estimated. However, since the transition between trunk and crown parts is smooth as trunks rejuvenate in crowns, the exact estimation of $h_t$ and $d_t$ remains challenging. This requires further investigations in future experiments. Our procedural model also confirms previous findings, such as optimal flying height and sampling distance, as well as the limits of AOS with respect to occlusion density. All of these findings suggest that AOS scanning with wide-FOV Nadir systems at high altitudes is counterproductive for occlusion removal.

For simplicity, the applied procedural tree model represent a statistical mixture of common conifer and broadleaf forests. In future, other procedural algorithms will be explored that are more specific to vegetation types which do not follow our occlusion distribution assumption (denser and smaller at the top, sparser and larger at the bottom).

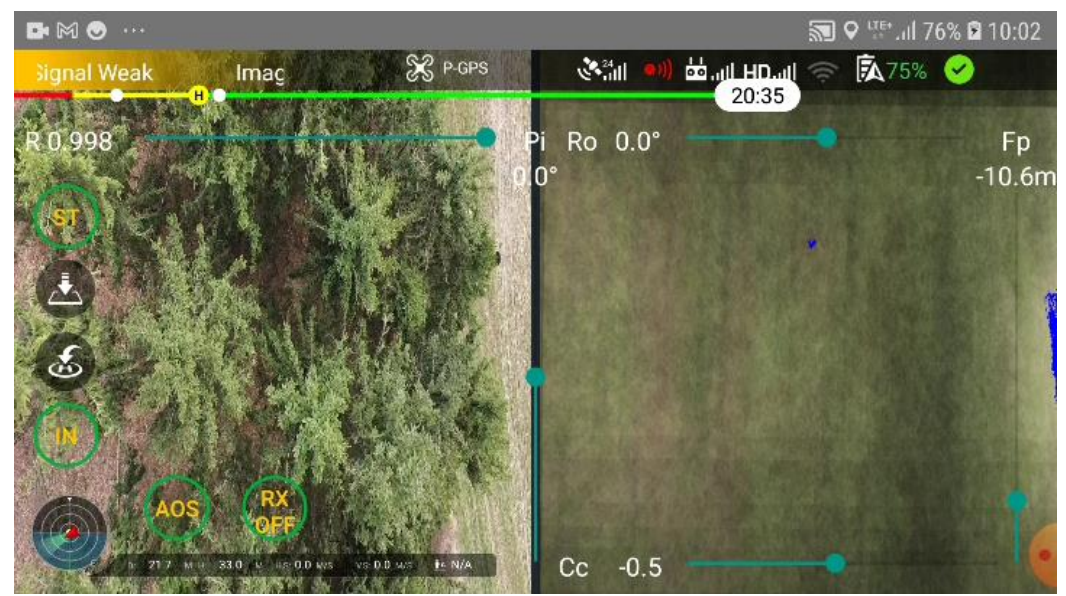

**Figure 7.** Freely available app that makes AOS freely available with DJI compatible platforms. Left screen half shows drone's live-video stream. Right screen half shows real-time AOS integration images with results of anomaly detection [36] (blue pixels in center) of a detected person. Available at: https://github.com/JKU-ICG/AOS/

While AOS has been explored with automatic and autonomous research prototypes in the past, we have published a freely available DJI compatible app that integrates AOS (cf. Fig. 7). This app enables blue light organizations and others to use and explore AOS with compatible, manually operated, off-the-shelf drones. In this implementation, we have chosen a default (digitally cropped) FOV of 50° for integral images that is based on our findings above (raw areal images are still shown at original camera FOV).

**Author Contributions:** Conceptualization and methodology, O.B.; software, validation, and formal analysis, F.S., I.K.; writing—original draft preparation O.B. and F.S.; writing—review, editing, and visualization, O.B., F.S., and I.K.; supervision, O.B.; project administration, O.B.; funding

acquisition, O.B; Implementation AOS DJI app, I.K, R.J.A.A.N., R.O. All authors have read and agreed to the published version of the manuscript.

**Funding:** This research was funded by the Austrian Science Fund (FWF) under grant number P 32185-NBL, and by the State of Upper Austria and the Austrian Federal Ministry of Education, Science and Research via the LIT–Linz Institute of Technology under grant number LIT-2019-8-SEE-114.

**Data Availability Statement:** All source code, data, publications, and the DJI app are free for non-commercial usage and can be accessed at: https://github.com/JKU-ICG/AOS/. An online executable version and the source code of the procedural forest simulation is available at: https://aos.tensorware.app/ and https://github.com/tensorware/aos-simulation . The source code of the procedural tree algorithm is available at: https://github.com/supereggbert/proctree.js

**Conflicts of Interest:** The authors declare no conflict of interest.

## References

1.  Moreira, A.; Prats-Iraola, P.; Younis, M.; Krieger, G.; Hajnsek, I.; Papathanassiou, K.P. A tutorial on synthetic aperture radar. *IEEE Geosci. Remote Sens. Mag.* **2013**, *1*, 6–43.
2.  Li, C.J.; Ling, H. Synthetic aperture radar imaging using a small consumer drone. *Proc. IEEE Int. Symp. Antennas Propag. UNSC/URSI Nat. Radio Sci. Meeting,* Jul. 2015; pp. 685–686.
3.  Rosen, P.A. *et al.*; Synthetic aperture radar interferometry. *Proc. IEEE.* **2000**, *88*, 333–382.
4.  Levanda, R.; Leshem, A. Synthetic aperture radio telescopes. *IEEE Signal Process. Mag.* **2009**, *27*, 14–29.
5.  Dravins, D.; Lagadec, T.; Nuñez, P.D. Optical aperture synthesis with electronically connected telescopes. *Nature Commun.* **2015**, *6*, 6852.
6.  Ralston, T.S.; Marks, D.L.; Carney, P.S.; Boppart, S.A. Interferometric synthetic aperture microscopy. *Nature Phys.* **2007**, *3*, 129–134.
7.  Hayes, M.P.; Gough, P.T. Synthetic aperture sonar: A review of current status. *IEEE J. Ocean. Eng.* **2009**, *34*, 207–224.
8.  Hansen, R.E. Introduction to synthetic aperture sonar. In Sonar Systems Edited, Rijeka, Croatia: Intech, 2011. [Online]. Available: http://www.intechopen.com/books/sonar-systems/introduction-to-synthetic-aperture-sonar.
9.  Jensen, J.A.; Nikolov, S.I.; Gammelmark, K.L.; Pedersen, M.H. Synthetic aperture ultrasound imaging. *Ultrasonics.* **2006**, *44*, e5–e15. [Online]. Available: http://www.sciencedirect.com/science/article/pii/S0041624X06003374.
10. Zhang, H.K.; Cheng, A.; Bottenus, N.; Guo, X.; Trahey, G.E.; Boctor, E.M. Synthetic tracked aperture ultrasound imaging: Design, simulation and experimental evaluation. *Proc. SPIE.* **2016**, *3*, Art. No. 027001. [Online]. Available: http://www.ncbi.nlm.nih.gov/pmc/PMC4824841.
11. Barber, Z.W.; Dahl, J.R. Synthetic aperture ladar imaging demonstrations and information at very low return levels. *Appl. Opt.* **2014**, *53*, 5531–5537.
12. Turbide, S.; Marchese, L.; Terroux, M.; Bergeron, A. Synthetic aperture lidar as a future tool for earth observation. *Proc. Int. Conf. Space Opt. (ICSO)* 2019, *10563*, Art. No. 105633V. https://doi.org/10.1117/12.2304256.
13. Vaish, V.; Wilburn, B.; Joshi, N.; Levoy, M. Using plane + parallax for calibrating dense camera arrays. *Proc. CVPR.* 2004, *1*, 1.
14. Vaish, V.; Levoy, M.; Szeliski, R.; Zitnick, C.L.; Kang, S.B. Reconstructing occluded surfaces using synthetic apertures: Stereo, focus and robust measures. *Proc. IEEE Comput. Soc. Conf. Comput. Vis. Pattern Recognit. (CVPR)* 2006, *2*, 2331–2338.
15. Zhang, H.; Jin, X.; Dai, Q. Synthetic aperture based on plenoptic camera for seeing through occlusions. *Advances in Multimedia Information Processing-PCM*, Hong. R.; Cheng, W.-H.; Yamasaki, T.; Wang, M.; Ngo, C.-W.; Eds. Cham, Switzerland: Springer, **2018**, 158–167.
16. Yang, T.; Ma, W.; Wang, S.; Li, J.; Yu, J.; Zhang, Y. Kinetic based real time synthetic aperture imaging through occlusion. *Multimedia Tools Appl.* **2016**, *75*, 6925–6943. https://doi.org/10.1007/s11042015-2618-1.
17. Joshi, N.; Avidan, S.; Matusik, W.; Kriegman, D.J. Synthetic aperture tracking: Tracking through occlusions. *Proc. IEEE Int. Conf. Comput. Vis,* Rio de Janeiro, Brazil, Oct.2007, pp.1–8.
18. Pei, Z. *et al.*, Occluded-object 3D reconstruction using camera array synthetic aperture imaging. *Sensors.* **2019**, *19*, 607.
19. Yang, T. *et al.*, All-in-focus synthetic aperture imaging. *Proc. Eur. Conf. Comput. Vis.* Fleet, D.; Pajdla, T.; Schiele, B.; Tuytelaars, T.; Cham, Eds., Switzerland: Springer, 2014, 1–15.
20. Pei, Z.; Zhang, Y.; Chen, X.; Yang, Y.-H. Synthetic aperture imaging using pixel labeling via energy minimization. *Pattern Recognit.* **2013**, *46*, 174–187.
21. Kurmi, I.; Schedl, D.C.; Bimber, O. Airborne optical sectioning. *J. Imaging* **2018**, *4*, 102. https://doi.org/10.3390/jimaging4080102.
22. Bimber, O.; Kurmi, I.; Schedl, D.C. Synthetic aperture imaging with drones. *IEEE Comput. Graph. Appl.* **2019**, *39*, 8–15.
23. Kurmi, I.; Schedl, D.C.; Bimber, O. A statistical view on synthetic aperture imaging for occlusion removal. *IEEE Sensors J.* **2019**, *19*, 9374–9383.
24. Kurmi, I.; Schedl, D.C.; Bimber, O. Thermal airborne optical sectioning. *Remote Sens.* **2019**, *11*, 1668.

25. Schedl, D.C.; Kurmi, I.; Bimber, O. Airborne optical sectioning for nesting observation. *Sci. Rep.* **2020**, *10*, 7254.
26. Kurmi, I.; Schedl, D.C.; Bimber, O. Fast Automatic Visibility Optimization for Thermal Synthetic Aperture Visualization. *IEEE Geosci. Remote Sens. Lett.* **2021**, *18*, 836–840.
27. Kurmi, I.; Schedl, D.C.; Bimber, O. Pose Error Reduction for Focus Enhancement in Thermal Synthetic Aperture Visualization. *IEEE Geosci. Remote. Sens. Lett.* **2021**, to be published. https://doi.org/10.1109/LGRS.2021.3051718.
28. Schedl, D.C.; Kurmi, I.; Bimber, O. Search and rescue with airborne optical sectioning. *Nat. Mach. Intell.* **2020**, *2*, 783–790. https://doi.org/10.1038/s42256-020-00261-3.
29. Schedl, D.C.; Kurmi, I.; Bimber, O. An autonomous drone for search and rescue in forests using airborne optical sectioning. *Sci. Robot.* **2021**, *6*, eabg1188.
30. Kurmi, I.; Schedl, D.C.; Bimber, O. Combined person classification with airborne optical sectioning. *Sci. Rep.* **2022**, *12*, 3804. https://doi.org/10.1038/s41598-022-07733-z
31. Nathan, R.J.A.A.; Kurmi, I.; Schedl, D.C.; Bimber, O. Through Foliage Tracking with Airborne Optical Sectioning. *arXiv* **2021**, arXiv:2111,06959v2.
32. Wierzbicki, D. Multi-Camera Imaging System for UAV Photogrammetry. *Sensors.* **2018**, *18*, 2433. https://doi.org/10.3390/s18082433.
33. Schifano, L.; Smeesters, L.; Berghmans, F.; Dewitte, S. Wide-Field-of-View Longwave Camera for the Characterization of the Earth's Outgoing Longwave Radiation. *Sensors.* **2021**, *21*, 4444. https://doi.org/10.3390/s21134444.
34. Brodie, K.L.; Bruder, B.L.; Slocum, R.K.; Spore, N.J. Simultaneous Mapping of Coastal Topography and Bathymetry From a Lightweight Multicamera UAS. *IEEE Transactions on Geoscience and Remote Sensing.* **2019**, *57*, 6844-6864. https://doi.org/10.1109/TGRS.2019.2909026.
35. Han, J. *et al.*; Image Motion of Remote Sensing Camera With Wide Field of View Over the Antarctic and Artic. *IEEE Journal of Selected Topics in Applied Earth Observations and Remote Sensing.* **2021**, *14*, 3475-3484. https://doi.org/10.1109/JSTARS.2021.3066626.
36. Reed, I.S; Yu, X. Adaptive Multiple-Band CFAR Detection of an Optical Pattern with Unknown Spectral Distribution. IEEE Transactions on Acoustics, Speech and Signal Processing, 1990, 38, 1760-1770. https://doi.org/10.1109/29.60107.